\documentclass{article} 
\usepackage[final]{colm2026_conference}

\usepackage{microtype}
\usepackage{hyperref}
\usepackage{url}
\usepackage{booktabs}

\usepackage{lineno}

\definecolor{darkblue}{rgb}{0, 0, 0.5}
\hypersetup{colorlinks=true, citecolor=darkblue, linkcolor=darkblue, urlcolor=darkblue}

\usepackage{inconsolata}
\usepackage{array}
\usepackage{booktabs}
\usepackage{multirow}
\usepackage{float}
\usepackage{stfloats}
\usepackage{graphicx}
\usepackage[dvipsnames]{xcolor}
\usepackage[most]{tcolorbox}
\usepackage{soul}
\usepackage{tikz}
\usetikzlibrary{calc}
\usetikzlibrary{decorations.pathmorphing}

\makeatletter

\newcommand{\defhighlighter}[3][]{%
  \tikzset{every highlighter/.style={color=#2, fill opacity=#3, #1}}%
}

\defhighlighter{yellow}{.5}

\newcommand{\highlight@DoHighlight}{
  \fill [ decoration = {random steps, amplitude=1pt, segment length=15pt}
        , outer sep = -15pt, inner sep = 0pt, decorate
        , every highlighter, this highlighter ]
        ($(begin highlight)+(0,8pt)$) rectangle ($(end highlight)+(0,-3pt)$) ;
}

\newcommand{\highlight@BeginHighlight}{
  \coordinate (begin highlight) at (0,0) ;
}

\newcommand{\highlight@EndHighlight}{
  \coordinate (end highlight) at (0,0) ;
}

\newdimen\highlight@previous
\newdimen\highlight@current

\DeclareRobustCommand*\highlight[1][]{%
  \tikzset{this highlighter/.style={#1}}%
  \SOUL@setup
  \def\SOUL@preamble{%
    \begin{tikzpicture}[overlay, remember picture]
      \highlight@BeginHighlight
      \highlight@EndHighlight
    \end{tikzpicture}%
  }%
  \def\SOUL@postamble{%
    \begin{tikzpicture}[overlay, remember picture]
      \highlight@EndHighlight
      \highlight@DoHighlight
    \end{tikzpicture}%
  }%
  \def\SOUL@everyhyphen{%
    \discretionary{%
      \SOUL@setkern\SOUL@hyphkern
      \SOUL@sethyphenchar
      \tikz[overlay, remember picture] \highlight@EndHighlight ;%
    }{%
    }{%
      \SOUL@setkern\SOUL@charkern
    }%
  }%
  \def\SOUL@everyexhyphen##1{%
    \SOUL@setkern\SOUL@hyphkern
    \hbox{##1}%
    \discretionary{%
      \tikz[overlay, remember picture] \highlight@EndHighlight ;%
    }{%
    }{%
      \SOUL@setkern\SOUL@charkern
    }%
  }%
  \def\SOUL@everysyllable{%
    \begin{tikzpicture}[overlay, remember picture]
      \path let \p0 = (begin highlight), \p1 = (0,0) in \pgfextra
        \global\highlight@previous=\y0
        \global\highlight@current =\y1
      \endpgfextra (0,0) ;
      \ifdim\highlight@current < \highlight@previous
        \highlight@DoHighlight
        \highlight@BeginHighlight
      \fi
    \end{tikzpicture}%
    \the\SOUL@syllable
    \tikz[overlay, remember picture] \highlight@EndHighlight ;%
  }%
  \SOUL@
}
\makeatother

\title{
Extracting Arguments, Not Just Classifying Them: Instruction-Tuned LLMs for Generative Component Detection}

\author{Sofiane Elguendouze, Erwan Hain, Elena Cabrio \& Serena Villata\\
Université Côte d’Azur, CNRS, INRIA, I3S\\
Sophia-Antipolis, France \\
\texttt{\{name.lastname\}@univ-cotedazur.fr} \\
}

\begin{document}

\ifcolmsubmission
\linenumbers
\fi

\maketitle

\begin{abstract}
Argumentative component detection (ACD) is a core subtask of Argument(ation) Mining (AM) and one of its most challenging aspects, as it requires jointly delimiting argumentative spans and classifying them into components such as claims and premises. While research on this subtask remains relatively limited compared to other AM tasks, most existing approaches formulate it as a simplified sequence labeling problem, component classification, or a pipeline of component segmentation followed by classification. In this paper, we propose ITFACD, a novel approach based on instruction-tuned Large Language Models (LLMs) using compact instruction-based prompts, and reframe ACD as a language generation task, enabling arguments to be identified directly from plain text without relying on pre-segmented components. Experiments on standard benchmarks show that our approach achieves higher performance compared to state-of-the-art systems. To the best of our knowledge, this is one of the first attempts to fully model ACD as a generative task, highlighting the potential of instruction tuning for complex AM problems. Our code and the datasets used are openly available in the following \href{https://github.com/sofianeElguendouz/ITFACD}{GitHub repository}.
\end{abstract}

\section{Introduction}
\label{sec:intro}
Argument(ation) Mining (AM) \citep{10.1145/2850417, cabrio2018five, Stede2019, 10.1162/coli_a_00364} has emerged as a prominent research area within natural language processing, aiming to automatically identify and analyze the structure of arguments expressed in text. By decomposing argumentative discourse into meaningful components and relations, AM supports a wide range of downstream applications, including opinion analysis in large-scale deliberation, decision support, and educational technologies. Among the various subtasks in AM, Argumentative Component Detection (ACD), which consists of the identification and classification of argumentative components (ACs) such as claims and premises, remains one of the most challenging. This difficulty stems from the need to simultaneously delimit component boundaries and assign correct labels, often in the presence of implicit reasoning and loosely structured discourse. Most existing approaches to ACD continue to frame the task as a sequence labelling problem often relying on extensive feature engineering \citep{stab2014identifying}, multi-stage pipelines (component delimitation + component classification) \citep{morio-etal-2022-end} or a mere component type classification that depends on the availability of pre-segmented ACs \citep{chen-etal-2024-exploring-potential, cabessa-etal-2025-argument}. 

Recent advances in large language models (LLMs), particularly instruction-tuned models, have significantly reshaped the landscape of many NLP tasks. Their strong generative capabilities and sensitivity to task descriptions offer new opportunities for addressing complex prediction problems beyond traditional classification paradigms. It is therefore unsurprising that LLM-based approaches have recently been explored in AM \citep{gorur-etal-2025-large, cabessa-etal-2025-argument, favero2025leveragingsmallllmsargument}, primarily focusing on argument classification and argument relation extraction.

Despite leveraging large pre-trained models, ACD is still predominantly formulated as a sequence classification problem operating on pre-segmented (or pre-identified) argumentative units \citep{10.1007/978-3-031-70816-9_20, cabessa-etal-2025-argument}. While such formulations can yield competitive performance, they inherently assume the availability of accurately segmented argumentative spans, reducing the problem to component classification rather than jointly addressing boundary detection and labelling. This assumption is itself highly non-trivial as it constrains applicability in realistic settings where argumentative boundaries are not explicitly/easily marked. In addition, argumentative units are the basic building blocks upon which all high-level AM tasks rely. Without accurate component extraction, relation prediction, stance detection, argument graph construction or other tasks cannot be performed reliably and errors at the component extraction stage inevitably propagate to downstream modules.

In this work, we propose a novel perspective on ACD by reimagining it as a language generation task. Specifically, we introduce ITFACD (Instruction-tuning for Full ACD), a compact prompting approach for instruction-tuned LLMs, in which models are fine-tuned using concise, instruction-based prompts designed to guide both the delimitation (segmentation) and classification of ACs in a joint manner. We evaluate our approach on standard AM benchmarks and show that it achieves performance that surpasses state-of-the-art methods. Importantly, beyond competitive results, this work represents one of the early attempts to reconceptualize ACD as a generative task within an instruction-tuning paradigm. By modeling segmentation and classification jointly as a single text-to-structure generation process, we depart from conventional multi-stage pipelines that decompose the task into separate boundary detection and labeling steps. Our findings suggest that treating ACD as a unified generation problem not only simplifies the modeling framework but also offers a flexible and effective alternative to traditional formulations of AM subtasks, opening new directions for future research in the field.

\section{Related Work}

\subsection{Argument Mining Models}
A wide range of methods has been introduced for AM, initially exploiting manually designed syntactic and lexical features \citep{stab2014identifying}.
Subsequent work employed traditional supervised machine learning algorithms, such as Support Vector Machines \citep{stab2017, habernal-gurevych-2017-argumentation}. \cite{Menini_Cabrio_Tonelli_Villata_2018} proposed multi-class Support Vector Machine classifiers which achieved good results on the relation classification stage of the AM pipeline. One of the early learning-based approaches consisted of using more advanced neural network-based models including RNNs \citep{niculae-etal-2017-argument, potash-etal-2017-heres}.

Further studies predominantly adopted supervised learning paradigms based on auto-encoder transformer architectures such as BERT \citep{devlin-etal-2019-bert}, motivated by their strong ability to capture contextual information and long-range argumentative dependencies. For instance, \cite{goffredo:hal-03873412} tackled the task of fallacious argument classification in political debates, specifically those of the U.S. Presidential Campaigns. They first built a large corpus of political debates annotated with fallacious arguments. Then, they defined a transformer-based model architecture for fallacy classification, fine-tuned on argumentation features outperforming standard baselines. \cite{morio-etal-2022-end} introduced a multi-task training framework (MT-AM) that jointly learns from auxiliary corpora to improve performance across AM tasks. They designed an end-to-end model using a two-staged method (multi-task pre-training and target corpus fine-tuning). The approach allowed to integrate information across datasets by sharing a unified model backbone while maintaining corpus-specific output layers. Their system demonstrated that cross-corpus transfer can boost results, particularly for smaller target datasets. \cite{habernal2024mining} investigated AM in decisions of the European Court of Human Rights (ECHR), developing models built on pre-trained BERT and RoBERTa \citep{liu2019roberta}.

LLMs have become the dominant tools in NLP, showing competitive performance across multiple tasks. \cite{chen-etal-2024-exploring-potential} explored LLMs by focusing on “counter speech generation”. They assessed the performance of multiple LLMs on a broad set of computational argumentation tasks (argument mining and argument generation), under zero-shot and few-shot settings. Their findings demonstrated that LLMs surpassed baseline models on certain datasets. \cite{pojoni2023argument} proposed a method that first transcribes podcast episodes into text and then uses OpenAI’s GPT-4 (via ChatGPT) to extract ACs (main claim, premise, counterargument and rebuttal). The definition of an argument unit was however different from that in the literature, where it was considered as a statement synthesized from a direct quotation from the text and not an extracted portion of the text. 

\subsection{Argumentative Component Detection}
\label{sec:acd}
\cite{goudas_2014} proposed a multi-stage framework for ACD in Greek social media texts, first classifying sentences as argumentative or non-argumentative, and then applying a Conditional Random Field (CRF) model to identify the exact claim and premise spans within argumentative sentences. \cite{stab2017} modeled ACD as a sequence labeling task and established strong baselines for identifying claims and premises in noisy online discussions including CRF and feature-rich models combining lexical, syntactic, discourse, and embedding features.
\cite{favero2025leveragingsmallllmsargument} applied small open-source LLMs (Qwen 2.5 7B, Llama 3.1 8B, and Gemma 2 9B) with few-shot prompting and fine-tuning for argument segmentation and classification as a multi-stage pipeline. Their experiments were conducted on a single corpus, PERSUADE 2.0, which consists of argumentative essays written by English high-school students. They shown that fine-tuned models consistently outperform few-shot prompting across both tasks highlighting the benefits of task-specific adaptation even for smaller LLMs. \cite{chen-etal-2024-exploring-potential} evaluated LLMs such as GPT 3.5, Flan T5, and LLaMA 2 in zero-shot and few-shot settings across multiple AM tasks including claim and evidence classification (treated separately), stance detection, argument generation, argument summarization, and counter-speech generation. Their results demonstrate that LLMs achieve promising performance across these tasks. However, as illustrated in the prompt formulations provided in their study, the tasks related to argument detection are operationalized primarily as classification problems rather than as extraction tasks where argumentative units are assumed to be pre-segmented. Their approach remains dependent on the prior availability of these segmented argumentative units and sidestep the more challenging problem of jointly identifying and delimiting ACs directly from raw text. The work by \cite{cabessa-etal-2025-argument} also explored several LLMs reporting state-of-the-art performance across multiple AM benchmarks. Their study demonstrated effectiveness for tasks such as argumentative component classification, argument relation identification, and relation type classification. However, their focus remains limited to classification-based formulations as previously mentioned without argument boundary detection.

Among generative approaches to AM, the work of \cite{kawarada-etal-2024-argument} is, to the best of our knowledge, the closest to ours. Using FLAN-T5-XXL encoder-decoder model, the authors formulate AM as a text-to-text generation problem and jointly address ACD and relation prediction within a unified sequence-to-sequence framework. In contrast, our work focuses exclusively on ACD and frames it as an instruction-tuning task for decoder-only LLMs. This difference extends beyond model architecture, the learning paradigm is different. While they rely on conventional supervised text-to-text fine-tuning, we cast ACD as an instruction-based generation task in which models learn to augment raw text with argumentative annotations through natural-language instructions. A second key distinction lies in the task formulation itself. Rather than jointly optimizing multiple AM objectives, we isolate ACD from other AM sub-tasks, enabling the model to focus solely on argument delimitation and classification. This choice is motivated by the observation that many downstream AM tasks, including stance and relation prediction, assume the prior availability of ACs. Furthermore, although multi-task learning can improve generalization in some settings, it may also introduce task interference when objectives are not fully aligned, leading to what is known as negative transfer \cite{9392366}. In the context of AM, coupling ACD with relation prediction may exacerbate discrepancies in token-level generation because the model must simultaneously reconcile component identification with relational dependencies at discourse level, potentially diverting modeling capacity away from the core challenge of accurately detecting and delimiting ACs.

\section{Methodology}
\subsection{Task formalization}
As outlined in Section \ref{sec:intro}, we focus exclusively on the task of ACD (Argumentative Component Detection), one of the core challenges in AM. Unlike joint end-to-end frameworks that simultaneously address component detection and relation prediction \citep{eger2017neural}, we deliberately narrow our scope to ACD as recent state-of-the-art results indicate that argument relation classification has reached relatively high levels of effectiveness. The accurate identification and delimitation of ACs from raw text, in contrast, remains insufficiently explored and continues to present substantial challenges as ACs represent the building blocks upon which all high-level AM tasks rely. In this line, we aim to perform argumentative unit segmentation and classification within a unified framework. As discussed in Section \ref{sec:acd}, many state-of-the-art approaches reduce ACD to a classification task applied to pre-segmented argumentative units, a formulation that simplifies the problem but assumes the prior availability of accurately segmented spans, something that rarely holds in real-world scenarios. 

Our approach seeks to identify ACs directly from plain, unsegmented text. We consider two types of components: \textit{claims} and \textit{premises}. A \textit{claim} is a statement asserting a position, opinion, or proposition that can be supported or contested, often in relation to a debatable topic. A \textit{premise} is a supporting or opposing statement that provides justification, evidence, or reasoning for a claim (or another premise), including statistics, expert testimony, factual information, anecdotes, or illustrative examples. In the example below, the segment highlighted in orange represents a claim, the one in blue corresponds to a premise supporting that claim:
{\begin{center}
    `` ... \highlight[red]{the government should try to preserve minority languages}. This is because \highlight[cyan]{language can be seen as much more than just one method of communication} ... "
\end{center}

\subsection{Datasets}
\label{sec:data}
Three datasets have been used in our experiments: 
\textbf{(1) \texttt{USElecDeb60To16 - ED}} \citep{ijcai2019p944}: This dataset consists of annotated transcripts from televised U.S. presidential election debates spanning the period 1960–2016. The corpus captures spontaneous, spoken political discourse characterized by interruptions, rhetorical strategies, and implicit argumentation. Its dialogical structure and conversational nature make argumentative component detection particularly challenging, as argumentative boundaries are often less explicit. Furthermore it is considered as the largest argument-annotated dataset to date.
\textbf{(2) \texttt{Persuasive Essays - PE}} \citep{stab2017}: Initially introduced in \cite{stab-gurevych-2014-annotating}, this dataset has been later extended to finally contain 402 English essays collected from \href{essayforum.com}{essayforum.com}, an online platform where users seek feedback on written compositions such as essays and research papers. Compared to USElecDeb60To16, this corpus exhibit clearer argumentative organization and more explicit discourse markers, making it one of the most widely adopted benchmarks for ACD. \textbf{(3) \texttt{Web Discourse - WD}} \citep{habernal-gurevych-2017-argumentation}: This dataset comprises user-generated web texts addressing six controversial education-related topics. Unlike the relatively well-structured persuasive essays, this corpus reflects informal online discourse, including noisy language, heterogeneous writing styles, and loosely structured argumentation.

Importantly, these datasets vary substantially in their sizes, writing style, structural clarity and discourse complexity. This diversity exposes models to different linguistic phenomena and levels of argumentative explicitness, thereby equipping our approach with a more robust evaluation setting and allowing us to assess its generalization ability across textual genres with variable argumentation patterns. Table \ref{tab:datasets} provides statistics on these datasets (before the train-dev-test splitting). Further details can be found in the appendix in Table \ref{tab:comp_data_detailed}.

\begin{table}[ht]
\centering
\resizebox{.75\textwidth}{!}{%
\begin{tabular}{ccc}
\toprule
    \textbf{Dataset} & \multicolumn{2}{c}{\textbf{Argument Components}} \\ 
    \multirow{2}{*}{} & \multicolumn{1}{c}{Claim}    & Premise \\ \midrule
    USElecDeb60To16 - ED \citep{ijcai2019p944}& \multicolumn{1}{c}{29k} & 26k \\ 
    Persuasive Essays - PE \citep{stab2017}& \multicolumn{1}{c}{2257} & 3832 \\ 
    Web Discourse - WD \citep{habernal-gurevych-2017-argumentation}& \multicolumn{1}{c}{195} & 538 \\ 
    Merge & \multicolumn{1}{c}{31.5k} & 30.3k \\ 
    \bottomrule
\end{tabular}}
\caption{Datasets statistics.}
\label{tab:datasets}
\end{table}

\subsection{Method}
We reconceptualize the ACD process as a text generation problem where LLMs are instruction-tuned to reproduce the original plain text while inserting argumentative tags that explicitly demarcate the boundaries and nature of ACs. 

This is achieved through a standardized prompting template that specifies the detailed task description, the input text, and the format expected for the output. The prompts are constructed from the annotated datasets described above. To align the data with the generative formulation of the task, all corpora originally annotated using the BIO-tagging scheme\footnote{BIO (Beginning–Inside–Outside) is a common labeling scheme in sequence tagging tasks used to indicate the boundaries of structured elements within text.} are converted into XML-tagged text, where ACs are explicitly marked using tags such as \texttt{<premise>...</premise>} and \texttt{<claim>...</claim>} (see Figure \ref{fig:eg_prompt_answer} in the appendix).

For each data instance, we create input-output pairs consisting of the task description accompanied with the instructions and the original plain text, and its annotated version augmented with XML-style argumentative tags. Long documents are segmented into contiguous chunks of up to 1024 tokens, ensuring that ACs never span multiple chunks. If a potential boundary would split a claim/premise, the chunk is shortened and the component moved to the next chunk. In all datasets, ACs are substantially shorter than 1024 tokens, so truncation or fragmentation doesn't occur. During inference, each chunk is processed independently and the generated outputs are concatenated back in their original order to reconstruct the final document-level prediction. We suggest that chunking has a minimal impact on ACD performance. Each chunk typically contains multiple ACs (7.5 on average in the PE dataset) along with sufficient surrounding context, allowing the model to capture the local argumentative structure needed for ACD. Wider context might help for tasks requiring longer dependencies like argument relation detection.

\subsection{Models}
\label{sec:am_models}
The goal of this work is to evaluate the effectiveness of framing ACD as a generative instruction-tuning task rather than to benchmark the latest generation of LLMs. We therefore focus on a set of open-weight, text-only models that remain practical to fine-tune, deploy, and reproduce.
The selected models span different architectural families and parameter scales: GPT-2-XL-1.5B \citep{radford2019language}, OPT-6.7B \citep{zhang2022opt} and Llama-3-8B-Instruct \citep{grattafiori2024llama3herdmodels}. 

We deliberately avoid larger, multimodal, agentic, or proprietary systems (e.g. GPT-5.5, Qwen-3.6, gemma-4). In such models, performance gains may be driven by scale, multimodal pretraining, specialized reasoning objectives, or proprietary training data rather than by the proposed ACD formulation itself, making the contribution of the instruction-tuning framework harder to isolate. Relatively compact and open-weight models additionally ensure full reproducibility/transparency and enable controlled fine-tuning and architectural inspection, which is not possible with closed APIs where comparisons would be restricted to prompting-based settings. They also facilitate broader adoption in real-world settings, enabling deployment in computationally constrained environments, where cost and privacy considerations may limit the use of larger or closed models.

For comparison, we also include two encoder-based transformer architectures, RoBERTa and DeBERTa \citep{he2021debertadecodingenhancedbertdisentangled}, which are widely used in sequence labeling and classification tasks. Unlike the generative models, these baselines are trained under traditional token-level classification objective.

\subsection{Evaluation}
\label{sec:eval}
We developed a dedicated evaluation protocol for ACD in which both the reference annotations and predictions (tagged text generated by the models) are tokenized, converted into BIO scheme then aligned position-wise. We adopted a (i) strict alignment procedure to compute Macro-F1 scores aiming to evaluate the models' ability to augment plain text with argumentative annotations while remaining faithful to the original text. To this end, only tokens with identical surface forms (exact matching) in both sequences are evaluated. In fact, models introduce in rare cases text alterations such as new word insertions, modifications (spelling variations, synonym substitutions, poly-lexical replacements, etc.) or deletions, despite explicit instructions to replicate the input text without any alteration. Such deviations disrupt token alignment and impact the evaluation scores. The generated tokens can no longer be perfectly aligned with the original text even when the underlying argumentative structure is correctly identified. The affected misaligned and subsequent tokens are therefore excluded from evaluation. 

While decoding constraints (low temperature, restricted top-p) helped to limit this behavior, our evaluation has been extended with two additional fidelity heuristics that enable to identify alteration impact on the overall conclusions (see Table \ref{tab:three_eval_protocols_with_seed_variance}): (ii) a sequence-alignment-based evaluation that recovers matching regions outside local text alterations, and (iii) a fidelity-aware evaluation that recovers benign alterations that do not affect the semantic content. These alterations are grouped under several categories (morphological variations, tokenization artifacts, lexical substitutions, spelling variations, and insertions/deletions). Remaining mismatches, including occasional generation loops or unrelated substitutions, are grouped under the category ``Other errors" (see Table \ref{tab:text_alterations}).

\section{Experimental settings}
For the LLM-based experiments, decoding is configured to minimize stochasticity during generation. Specifically, we employ a very low temperature (0.01) combined with a restrictive nucleus sampling parameter (top-p = 0.1), encouraging near-deterministic outputs. To accommodate context window constraints, longer inputs are segmented into chunks of up to 1024 tokens. The models were fine-tuned using the standard causal language modeling objective, where the target sequence (the tagged text to generate) is conditioned on both the instruction prompt and the original unannotated text. Each model is fine-tuned for 20 epochs, and the checkpoint achieving the highest Macro-F1 score on the development set is selected for final evaluation. All experiments follow a standard 80/10/10 train–dev–test split and took place on A100 and H100 (80GB) GPUs.

For the encoder-based baselines, ACD is formulated as a token-level sequence labeling task using the conventional BIO scheme, where models are fine-tuned for token classification over 20 epochs. All models were hyperparameter-tuned\footnote{Hyperparameter-tunning using the \href{https://github.com/crscardellino/argumentation-mining-transformers}{Argumentation Mining Transformers Module (AMTM)}} over the following search space (the selected values are shown in bold): \texttt{batch\_size} $\in \{8, \mathbf{16}, 32\}$, \texttt{maximum\_sequence\_length} $\in \{\mathbf{64}, 128, 256, 512\}$ and \texttt{learning\_rate} $\in \{\mathbf{1e-4}, 2e-4, 3e-4, 4e-4\}$. Although they technically accept a larger context window, feeding larger context (e.g. 512 tokens) to this models does not guarantee better use of long-range information. In practice, performance tends to plateau once the window is large enough to capture surrounding sentence or paragraph that provides sufficient context for accurate classification, along with a good trade-off between \texttt{batch\_size} and \texttt{maximum\_sequence\_length} which generally guarantees for such models a more stable gradient.

Concerning the fidelity statistics and fidelity-aware heuristics discussed in Section \ref{sec:eval}, the spelling variations were detected using fuzzy matching with a \texttt{similarity\_threshold} $\ge 80$, lexical substitutions using WordNet, morphological variations through POS-aware lemmatization (spaCy), insertions/deletions through local token alignment, and tokenization artifacts through several text normalizations.

All experiments on the \texttt{PE} dataset are conducted across five independent random seeds $\{10, 20, 30, 40, 50\}$ and results are reported as mean $\pm$ standard deviation. This provides a more reliable estimate of performance and ensures that the reported gains are not attributable to a favorable single run.

\section{Results and discussion}
\subsection{Instruction tuning setting}
Table \ref{tab:acd-results} reports the performance of our models on the ACD task in terms of Macro-F1 score, widely considered the standard evaluation metric in AM. We first present results across the five seeds on the \texttt{PE} benchmark, the most widely used benchmark for ACD, to enable direct comparison with established baselines. We then report performance on the individual datasets \texttt{ED} and \texttt{WD}, followed by the \texttt{Merge} setting, which combines all three corpora \texttt{(PE, ED, WD)} and provides a more challenging cross-domain evaluation scenario with heterogeneous writing styles and discourse complexity, as previously outlined in Section \ref{sec:data}.

On the \texttt{PE} dataset, the Llama-3 model achieved the best overall performance with an average Macro-F1 score of 0.8931, outperforming every reported baseline, including the strongest feature-engineered CRF model \citep{stab2017}. It is worth highlighting that the latter relies on syntactic parsing and extensive task-specific hand-crafted features, whereas our instruction-tuned models learn directly through generative supervision. Interestingly, the obtained score slightly exceeds the reported human upper bound of 0.8860 \citep{stab2017}. This result should not be interpreted as the model being "better than human annotators"; rather, the reported upper bound reflects inter-annotator agreement measured under a particular annotation protocol and evaluation setting. Since the model is trained to reproduce the annotations of a fixed gold standard, it can achieve a level of agreement with that reference that is comparable to, or even marginally higher than, the agreement observed between individual human annotators. Nevertheless, the small difference ($0.0071$) suggests that the proposed approach reaches a level of consistency with the reference annotations that is close to the limits imposed by annotation variability in the dataset. In Table \ref{tab:class_wise_acd_pe} in the Appendix, we provide the BIO-class-wise Macro-F1 scores on the PE dataset, for a more in-depth analysis.

Beyond the \texttt{PE} benchmark, our models exhibit distinct behaviors across datasets, providing further insight into their generalization capabilities. On the \texttt{ED} dataset, which consists of dialogical and relatively unstructured political debate transcripts where argumentative boundaries are less explicit and often embedded in longer, context-dependent exchanges, all models show a performance drop compared to the \texttt{PE} dataset.

\begin{table*}[ht]
    \centering
    \resizebox{\textwidth}{!}{%
    \begin{tabular}{lccccc}
        & \textbf{Params} & \textbf{PE} & \textbf{ED} & \textbf{WD} & \textbf{Merge}  \\
        \cmidrule(lr){2-6}
        Human Upper Bound \citep{stab2017} & - & \textbf{0.8860} & & & \\
        \midrule
        Heuristic Baseline\citep{stab2017} & - & 0.6420 & - & - & - \\
        CRF with features \citep{stab2017} & - & 0.8670 & - & - & - \\
        MT-all \citep{morio-etal-2022-end} & - & 0.7566 & - & - & - \\
        FLAN T5-XXL \citep{kawarada-etal-2024-argument} & 11B & 0.8015 & - & - & - \\
        \midrule
        ITFACD-OPT & 6.7B & 0.8233 $\pm$ 0.0104 & \textbf{0.7600} & 0.4625 & \textbf{0.7822}  \\
        ITFACD-GPT-2 & 1.5B & 0.8419 $\pm$ 0.0075 & 0.7476 & 0.4785 & 0.7684  \\
        ITFACD-Llama-3 & 8B & \textbf{0.8931} $\pm$ 0.0059 & 0.7091 & \textbf{0.5615} & 0.7667  \\
        DeBERTa-v3 & 86M & 0.7094 $\pm$ 0.0090 & 0.4730 & 0.3209 & 0.4913 \\
        RoBERTa & 125M & 0.6885 $\pm$ 0.0098 & 0.4741 & 0.3671 & 0.4844 \\
        \bottomrule
    \end{tabular}}
    \caption{ACD model assessment (Strict Macro-F1).}
    \label{tab:acd-results}
\end{table*}

On the more challenging \texttt{WD} dataset, which contains noisy and informal web discourse, Llama-3 achieves the best result, outperforming all other models by a noticeable margin. All models exhibit lower performance compared to every setting, potentially due to their sensitivity to linguistic variability (inconsistent syntax, implicit argumentation, heterogeneous writing styles etc.), and their lower capability of generalizing to highly informal text.

In addition to the single-dataset settings, the \texttt{Merge} corpus introduces a challenging and insightful scenario, as models are required to learn/generalize across heterogeneous domains simultaneously. Although performance is slightly below the best single-domain results obtained on the \texttt{PE} dataset, it remains substantially higher than on (the more challenging) \texttt{WD} and, to a lesser extent, \texttt{ED}. Combining datasets exposes the models to a broader range of writing styles, discourse structures, and argumentative patterns, enabling them to learn more robust and transferable representations, which appears to mitigate the difficulty of noisy and informal data present in some single domain corpora. Overall, this highlights the advantage of multi-domain training for ACD \citep{schiller-etal-2024-diversity}, as it encourages models to move beyond dataset-specific regularities and develop a more generalized understanding of argumentative structure.

Encoder-based transformer baselines (RoBERTa and DeBERTa-v3) trained on \texttt{PE} achieve good results and surpass the earlier heuristic baseline. However, both models remain below the performance of the LLM models. These models experience a much sharper degradation in the other settings, suggesting that traditional encoder-based models struggle more with capturing cross-domain variability and the variability of argumentative structures across heterogeneous corpora than generative models.

The superiority of the generative formulation can be attributed to several factors. First, the decoder-only architecture inherently model longer-range dependencies through autoregressive attention, enabling them to capture discourse-level cues and implicit argumentative structures that may span multiple sentences. This is particularly advantageous for ACD, where argumentative boundaries are often not signaled by explicit markers and may depend on precise contextual interpretation. Second, by casting the task as structured text generation with explicit XML-style tags, the model learns to internalize both boundary detection and labeling jointly, rather than propagating errors across separate pipeline stages. This reduces cumulative segmentation–classification error and encourages globally coherent predictions. Our work is further complemented by a qualitative analysis (Section \ref{sec:qualit_study} in the Appendix), which highlights potential strengths of the generative formulation. In several cases, the models correct annotation inconsistencies and identify plausible yet unannotated ACs, suggesting a capacity to capture subtle argumentative structures.

\subsection{Zero/One-shot setting}

\begin{table*}[bp]
\centering
\resizebox{\textwidth}{!}{%
\begin{tabular}{lllcccccccc}
\toprule
\multirow{2}{*}{\textbf{Model}} & \multirow{2}{*}{\textbf{Params}} & \multirow{2}{*}{\textbf{Learning}} & \multicolumn{2}{c}{\textbf{PE}} & \multicolumn{2}{c}{\textbf{ED}} & \multicolumn{2}{c}{\textbf{WD}} & \multicolumn{2}{c}{\textbf{Merge}} \\
\cmidrule(lr){4-5}
\cmidrule(lr){6-7}
\cmidrule(lr){8-9}
\cmidrule(lr){10-11}
& & & \textbf{Strict} & \textbf{Seq. Align.} & \textbf{Strict} & \textbf{Seq. Align.} & \textbf{Strict} & \textbf{Seq. Align.} & \textbf{Strict} & \textbf{Seq. Align.} \\
\midrule

\multirow{2}{*}{GPT-2} & \multirow{2}{*}{1.5B} & Z/s & 0.0968 & 0.1710 (12.99\%) & 0.1431 & 0.1971 (18.50\%) & 0.1523 & 0.1595 (16.36\%) & 0.1263 & 0.1712 (16.04\%) \\
& & O/s & 0.1836 & 0.1762 (48.10\%) & 0.1409 & 0.1745 (24.42\%) & 0.1409 & 0.1708 (39.37\%) & 0.1304 & 0.1583 (26.33\%) \\
\midrule

\multirow{2}{*}{OPT} & \multirow{2}{*}{6.7B} & Z/s & 0.1097 & 0.1096 (39.67\%) & 0.0873 & 0.1084 (20.39\%) & 0.1486 & 0.1893 (25.80\%) & 0.1042 & 0.1255 (19.43\%) \\
& & O/s & 0.1162 & 0.1103 (60.34\%) & 0.1297 & 0.1243 (6.90\%) & 0.1565 & 0.1540 (35.05\%) & 0.1442 & 0.1313 (3.04\%) \\
\midrule

\multirow{2}{*}{Llama-3} & \multirow{2}{*}{8B} & Z/s & 0.1130 & 0.1382 (32.90\%) & 0.1455 & 0.1909 (21.76\%) & 0.1589 & 0.1645 (22.32\%) & 0.1265 & 0.1344 (20.10\%) \\
& & O/s & 0.2049 & 0.2292 (55.10\%) & 0.1523 & 0.1961 (26.01\%) & 0.1577 & 0.1879 (38.12\%) & 0.1782 & 0.1804 (28.33\%) \\
\bottomrule
\end{tabular}}
\caption{Zero-shot (Z/s) and one-shot (O/s) prompting results under strict and sequence-alignment evaluation protocols. Values in parentheses indicate text preservation rates.}
\label{tab:zero_one_shot_results}
\end{table*}

As no literature baselines report full ACD results on the \texttt{ED} and \texttt{WD} datasets (and consequently on Merge), we conducted additional zero/one-shot prompting experiments. The results are reported in Table \ref{tab:zero_one_shot_results}. The main conclusion that can be drawn here is that prompting alone remains challenging for ACD, with Macro-F1 generally below 0.23 and substantially lower than their fine-tuned counterparts, highlighting the importance of task-specific fine-tuning. Two additional trends emerge: one-shot prompting usually improves performance over zero-shot, and better ACD performance is generally associated with higher preservation rates (reported in parentheses and measuring the extent to which the generated output faithfully reproduces the input text while only inserting argument tags). The often large gap between strict and sequence-alignment evaluation further indicates that a significant portion of the performance loss in zero/one-shot settings originates from text replication issues. In contrast, this gap is much smaller with instruction-tuning (Table \ref{tab:three_eval_protocols_with_seed_variance}), demonstrating its ability to improve both ACs detection and adherence to the required output format.

\subsection{Fidelity analysis}
Table \ref{tab:text_alterations} reports fidelity statistics on the \texttt{PE} dataset across five random seeds. Text preservation is consistently high across all models, with discrepancies affecting only a small fraction of generated tokens and distributed across the six categories described in Section \ref{sec:eval}. Text alterations remain rare for OPT and GPT-2, with average alteration rates of only 0.89\% and 0.23\% (resp.). The rate is higher for Llama-3 (7.2\%); however, most cases correspond to generation failures rather than genuine lexical modifications, including decoding artifacts and difficulties in consistently producing well-formed XML tags. Further investigation identified several decoding-related causes. In particular, suboptimal setting of \texttt{max\_new\_tokens} occasionally caused premature termination, \texttt{no\_repeat\_ngram\_size} sometimes triggered generation loops, and \texttt{top\_k=0} left the model's vocabulary unconstrained during decoding, increasing the likelihood of undesirable lexical substitutions.

Table \ref{tab:three_eval_protocols_with_seed_variance} presents models' mean Macro-F1 $\pm$ standard deviation across the five seeds on the \texttt{PE} dataset under the three evaluation protocols discussed in Section \ref{sec:eval}. Overall, variance is low for all models (std $< 0.011$) indicating stable and reproducible performance. For OPT and GPT-2, scores are nearly identical across evaluation protocols, confirming that text alterations have a negligible impact on the reported scores. Llama-3 exhibits the lowest std of all models, making it the most stable model across runs. Although its scores decrease under the sequence-alignment and fidelity-aware evaluations, this behavior is consistent with the fidelity analysis reported above and primarily reflects generation-formatting issues rather than failures in identifying ACs. Llama-3 remains the best-performing model under all evaluation protocols, and outperforms the strongest baseline reported in Table \ref{tab:acd-results}.

\begin{table}[tp]
\centering
\small
\resizebox{\columnwidth}{!}{%
\begin{tabular}{lccccccc}
\toprule
\textbf{Model} & \textbf{Exact} & \textbf{Morpho.} & \textbf{Token.} & \textbf{Lexical} & \textbf{Spelling} & \textbf{Ins./Del.} & \textbf{Other} \\
 & \textbf{preserv.} & \textbf{var.} & \textbf{artifacts} & \textbf{substit.} & \textbf{var.} &  & \textbf{errors} \\
\midrule
ITFACD-OPT & 99.11\% $\pm$ 1.04\% & 0.04\% $\pm$ 0.03\% & 0.03\% $\pm$ 0.02\% & 0.03\% $\pm$ 0.01\% & 0.03\% $\pm$ 0.01\% & 0.07\% $\pm$ 0.06\% & 0.69\% $\pm$ 0.92\% \\

ITFACD-GPT-2 & 99.77\% $\pm$ 0.18\% & 0.03\% $\pm$ 0.01\% & 0.01\% $\pm$ 0.00\% & 0.02\% $\pm$ 0.01\% & 0.02\% $\pm$ 0.01\% & 0.03\% $\pm$ 0.01\% & 0.13\% $\pm$ 0.15\% \\

ITFACD-Llama3 & 92.10\% $\pm$ 0.61\% & 0.04\% $\pm$ 0.01\% & 0.51\% $\pm$ 0.02\% & 0.28\% $\pm$ 0.02\% & 0.11\% $\pm$ 0.02\% & 1.37\% $\pm$ 0.04\% & 5.59\% $\pm$ 0.53\% \\
\bottomrule
\end{tabular}
}
\caption{Distribution of text preservation and alteration categories across models.}
\label{tab:text_alterations}
\end{table}

\begin{table}[tp]
\centering
\scriptsize
\begin{tabular}{lcccc}
\toprule
\textbf{Model} & \textbf{Params} & \textbf{Strict} & \textbf{Seq. Align.} & \textbf{Fidelity-Aware} \\
\midrule
ITFACD-OPT & 6.7B & $0.8233 \pm 0.0104$ & $0.8233 \pm 0.0106$ & $0.8228 \pm 0.0105$ \\
ITFACD-GPT-2 & 1.5B & $0.8419 \pm 0.0075$ & $0.8406 \pm 0.0069$ & $0.8409 \pm 0.0069$ \\
ITFACD-Llama-3 & 8B & $\mathbf{0.8931} \pm 0.0059$ & $\mathbf{0.8451} \pm 0.0055$ & $\mathbf{0.8471} \pm 0.0073$ \\
\bottomrule
\end{tabular}
\caption{ACD model assessment under the three evaluation protocols (on the \texttt{PE} dataset).}
\label{tab:three_eval_protocols_with_seed_variance}
\end{table}

\section{Conclusion}
In this work, we revisited Argumentative Component Detection (ACD) in Argumentation Mining (AM) through the lens of instruction-tuned Large Language Models (LLMs). Departing from traditional sequence labeling and multi-stage pipeline approaches, we reformulated ACD as a unified generative task that jointly performs argument delimitation and labeling directly from raw text. Our results demonstrate that instruction-tuned open-weight LLMs can effectively handle this structured prediction task. Our best performing model achieved a Macro-F1 score of 0.8931, surpassing all considered baselines. This work represents one of the early attempts to fully cast ACD as a generative task, opening promising directions for extending this paradigm to other AM subtasks.

\section*{Limitations}
Despite the encouraging results in this work, several limitations must be acknowledged. Hallucinations and generation fidelity continue to be important challenges for LLM-based approaches. In our setting, these issues primarily manifest as failures to exactly reproduce the input text, despite explicit instructions to preserve it and extensive decoding hyperparameter tuning. While decoding constraints (low temperature, restricted top-p) helped us to reduce this behavior, they did not eliminate them entirely. A promising direction for future work is to move from heuristic mitigation to constrained generation. It should be feasible to constrain generation so that the model can only emit either the next token from the source text or one of the valid annotation tags. Existing approaches provide promising foundations in this direction, including grammar-constrained generation methods that enforce user-defined constraints through finite-state control of the decoding process \cite{geng-etal-2023-grammar}, and more recent constrained discrete diffusion frameworks that impose hard sequence-level constraints during generation \cite{NEURIPS2025_124c1cc9}, which could strengthen both prompting-based and fine-tuned ACD systems.
Our experiments focus exclusively on claim and premise detection. While this choice was intentional to prioritize unified argument component extraction, other important argument mining subtasks such as relation identification, stance modeling, and full argument graph construction remain unexplored within the generative framework.

\section*{Ethics Statement}
The datasets used in this work might reflect domain/language-specificities, and argumentative annotations inherently involve subjective judgments. Consequently, the models may reproduce or amplify potential existing biases and should not be interpreted as providing objective or universally valid argument structures. Also, argument boundary identification is inherently interpretative, and generative models may expose annotation inconsistencies or implicitly reshape argumentative structure, potentially presenting subjective analyses as objective outputs. Moreover, as observed in our experiments, LLMs may hallucinate or slightly alter input text, which can distort meaning in sensitive contexts. We emphasize that the proposed system is intended for analytical and research purposes, as argumentation technologies may be misused in contexts such as political persuasion or large-scale rhetorical analysis, particularly when combined with generative capabilities.

\section*{Acknowledgments}
This work was granted access to the HPC resources of IDRIS under the allocation 2026-AD011016080R1 made by GENCI. It was partially supported by the French government (ANR, France 2030 investment plan) through the 3IA Côte d'Azur programme (ANR-23-IACL-0001), and by the European Union's project ORBIS under the Horizon Europe programme (Grant Agreement No. 101094765). The work of Serena Villata in this paper was also supported by ERC grant PANDORA-101231869, funded by the European Union. The views and opinions expressed are those of the authors only and do not necessarily reflect those of the European Union, the European Research Council Executive Agency, or any other granting authority. Neither the European Union nor the granting authorities can be held responsible for them.
 
\bibliography{bib.bib}
\bibliographystyle{colm2026_conference}

\newpage
\section*{Appendix}

\subsection*{Dataset statistics}
\begin{table}[ht]
\centering
\resizebox{.7\textwidth}{!}{%
\begin{tabular}{llllll}
\toprule
    \textbf{Dataset} & \textbf{O} & \textbf{B-Premise} & \textbf{I-Premise} & \textbf{B-Claim} & \textbf{I-Claim}\\\midrule
    USElecDeb60To16 & 566492 & 26055 & 350079 & 29624 & 338941 \\ 
    Persuasive Essays & 35946 & 2257 & 29828 & 3832 & 59652 \\ 
    Web Discourse & 61414 & 195 & 3491 & 538 & 20566 \\ \bottomrule
\end{tabular}}
\caption{Detailed statistics on the datasets used for fine-tuning and testing LLM models following the BIO-tagging scheme}
\label{tab:comp_data_detailed}
\end{table}

\subsection*{Class-wise ACD on the PE dataset}
A more fine-grained analysis of component detection performance under the BIO tagging scheme is provided in Table \ref{tab:class_wise_acd_pe}, which enables a closer examination of boundary detection accuracy and class-specific performance on the PE dataset.

Instruction-tuned LLMs clearly outperform BERT-based baselines across all argument classes. In particular, Llama-3-8B achieves the strongest performance, with balanced scores for both claim boundaries (B-Claim: 0.8513, I-Claim: 0.8258) and premise boundaries (B-Premise: 0.9096, I-Premise: 0.8997), resulting in the highest Macro-F1 (0.8931). Notably, premise components (B-Premise and I-Premise) are generally detected more accurately than claims across all generative models, suggesting that they exhibit more consistent lexical or structural cues making them easier to detect than claims. The O class consistently obtains the highest scores (up to 0.9789), which is expected given its higher frequency and clearer separation from argumentative spans. DeBERTa-v3 and RoBERTa show substantially lower performance on B- and I- tags, especially for claim beginnings, indicating difficulty in precise span delimitation.

\begin{table}[ht]
    \centering \scriptsize
    \resizebox{\textwidth}{!}{%
    \begin{tabular}{cccccccc}
    \toprule
        \textbf{Model} & \textbf{B-Claim} & \textbf{I-Claim} & \textbf{B-Premise} & \textbf{I-Premise} & \textbf{O} & \textbf{Macro-F1 (Strict)} \\\midrule
        OPT-6.7B & 0.7432 $\pm$ 0.0124 & 0.7099 $\pm$ 0.0109 & 0.8458 $\pm$ 0.0165 & 0.8546 $\pm$ 0.0140 & 0.9632 $\pm$ 0.0043 & 0.8233 $\pm$ 0.0104 \\
        GPT-2-1.5B & 0.7671 $\pm$ 0.0106 & 0.7426 $\pm$ 0.0104 & 0.8614 $\pm$ 0.0067 & 0.8732 $\pm$ 0.0068 & 0.9654 $\pm$ 0.0050 & 0.8419 $\pm$ 0.0075 \\
        Llama-3-8B & 0.8513 $\pm$ 0.0098 & 0.8258 $\pm$ 0.0092 & 0.9096 $\pm$ 0.0059 & 0.8997 $\pm$ 0.0067 & 0.9789 $\pm$ 0.0032 & 0.8931 $\pm$ 0.0059 \\
        DeBERTa-v3	& 0.6032 $\pm$ 0.0095 & 0.7367 $\pm$ 0.0101 & 0.5691 $\pm$ 0.0113 & 0.8013 $\pm$ 0.0098 & 0.8356 $\pm$ 0.0089 & 0.7094 $\pm$ 0.0110 \\
        RoBERTa & 0.6011 $\pm$ 0.0110 & 0.6787 $\pm$ 0.0079 & 0.5923 $\pm$ 0.0107 & 0.7714 $\pm$ 0.0065 & 0.7988 $\pm$ 0.0073 & 0.6885 $\pm$ 0.0098 \\\midrule
    \end{tabular}}
    \caption{Detailed Strict Macro-F1 scores for ACD on the \texttt{PE} dataset. B: Beginning, I: Inside, O: Outside.}
    \label{tab:class_wise_acd_pe}
\end{table}

\subsection*{Qualitative study}
\label{sec:qualit_study}
A human inspection of model outputs revealed several noteworthy qualitative phenomena that are not fully captured by standard automatic metrics. Beyond typical correct and incorrect label assignments (Table \ref{tab:errors-qualit-results}), we observed three recurring patterns that highlight both the strengths and limitations of the generative formulation.

The first phenomenon concerns what we refer to as argument type refinement. In several cases, the model (OPT-6.7B) predicts a component label that differs from the gold annotation but is arguably more appropriate given the context. In such instances, the generated label exposes potential inconsistencies or borderline cases in the original annotation. An example is illustrated in Tables \ref{tab:type-enhancement-and-rephrasing-qualit-results} and \ref{tab:type-enhancement-1-qualit-results}. Although these predictions are currently penalized as errors under strict token-level evaluation, they may reflect legitimate interpretative alternatives rather than genuine mis-labelling. This observation suggests that instruction-tuned LLMs can implicitly learn nuanced distinctions in argumentative structure and may even help identify annotation noise or ambiguities in existing corpora.

The second phenomenon involves the identification of previously unannotated ACs. In some cases, the model marks spans as claims or premises that were not labeled in the ground-truth data but should reasonably be interpreted as argumentative. This behavior demonstrates the model’s capacity to generalize argumentative patterns beyond the annotated instances and to detect implicit argumentative structures that may have been overlooked during manual annotation. While these predictions are again treated as false positives under the current evaluation framework, they highlight the model’s ability to capture deeper discourse-level reasoning. Illustrative examples are shown in Tables \ref{tab:discovery-qualit-results} and \ref{tab:discovery-1-qualit-results}.

The third phenomenon relates to surface-level textual changes. The models occasionally introduce minor grammatical or lexical adjustments during generation, such as correcting verb conjugations, modifying singular/plural forms, or slightly rephrasing segments of the text (see Table \ref{tab:type-enhancement-and-rephrasing-qualit-results}). Although these changes preserve both the semantic meaning and the argumentative role of the content, they result in exact string mismatches when compared token-by-token with the reference output. As our current evaluation protocol relies on strict span alignment, such discrepancies lead to these predictions being counted as errors, even when the underlying argument structure is correctly identified. This underscores a possible limitation arising both from the inherent variability of LLM generation and from the rigidity of the current evaluation protocol for our generative framework. It points to the need for future work to better control generation and mitigate non-deterministic behaviour, while also developing more flexible and reliable evaluation methods capable of capturing such variations.

\begin{table}[ht]
    \centering
    \scriptsize
    \resizebox{\textwidth}{!}{%
    \begin{tabular}{p{0.1\textwidth}p{0.7\textwidth}}
    \toprule
        \textbf{Gold} &  And so \textcolor{cyan}{\texttt{<premise>}}if you believe the same thing\textcolor{cyan}{\texttt{</premise>}}, \textcolor{red}{\texttt{<claim>}}you just don't want to raise taxes on people\textcolor{red}{\texttt{</claim>}}. And \textcolor{red}{\texttt{<claim>}}the reality is it's not just wealthy people\textcolor{red}{\texttt{</claim>}}\\
        \midrule
        \textbf{ITFACD-OPT-6.7B} & And so \textcolor{cyan}{\texttt{<premise>}}if you believe the same thing\textcolor{cyan}{\texttt{</premise>}}, \textcolor{red}{\texttt{<claim>}}you just don't want to raise\textcolor{red}{\texttt{</claim>}} taxes on people. And \textcolor{cyan}{\texttt{<premise>}}the reality is it's not just wealthy people\textcolor{cyan}{\texttt{</premise>}}\\
        \bottomrule
    \end{tabular}}
    \caption{Example of ACD with argument type errors}
    \label{tab:errors-qualit-results}
\end{table}

\begin{table}[ht]
    \centering
    \scriptsize
    \resizebox{\textwidth}{!}{%
    \begin{tabular}{p{0.1\textwidth}p{0.7\textwidth}}
    \toprule
        \textbf{Gold} & \textcolor{red}{\texttt{<claim>}}We will do what we do best\textcolor{red}{\texttt{</claim>}}. \textcolor{red}{\texttt{<claim>}}It's a strategy that we've been working on for a \textbf{couple of years}\textcolor{red}{\texttt{</claim>}}. \textcolor{red}{\texttt{<claim>}}It is going to take us to much better advantage in conventional forces\textcolor{red}{\texttt{</claim>}}\\
        \midrule
        \textbf{ITFACD-OPT-6.7B} & \textcolor{red}{\texttt{<claim>}}We will do what we did best\textcolor{red}{\texttt{</claim>}}. \textcolor{cyan}{\texttt{<premise>}}It's a strategy that we've been working on for a \textbf{few years}\textcolor{cyan}{\texttt{</premise>}}. \textcolor{red}{\texttt{<claim>}}It is going to take us to much better advantage in conventional forces\textcolor{red}{\texttt{</claim>}}\\
        \bottomrule
    \end{tabular}}
    \caption{Example of ACD with argument type refinement and lexical adjustment}
    \label{tab:type-enhancement-and-rephrasing-qualit-results}
\end{table}

\begin{table}[ht]
    \centering
    \scriptsize
    \resizebox{\textwidth}{!}{%
    \begin{tabular}{p{0.1\textwidth}p{0.7\textwidth}}
    \toprule
        \textbf{Gold} & \textcolor{red}{\texttt{<claim>}}Maybe we need to do a better job in mental clinics to help them\textcolor{red}{\texttt{</claim>}}. Because \textcolor{cyan}{\texttt{<premise>}}there is a major problem there\textcolor{cyan}{\texttt{</premise>}}\\
        \midrule
        
        \textbf{ITFACD-OPT-6.7B} & \textcolor{red}{\texttt{<claim>}}Maybe we need to do better job in mental clinics\textcolor{red}{\texttt{</claim>}} \textbf{\textcolor{cyan}{\texttt{<premise>}}to help them\textcolor{cyan}{\texttt{</premise>}}}. Because \textcolor{cyan}{\texttt{<premise>}}there is a major problem there\textcolor{cyan}{\texttt{</premise>}}.\\
        \bottomrule
    \end{tabular}}
    \caption{Example of ACD with argument components discovery}
    \label{tab:discovery-qualit-results}
\end{table}

\begin{table}[ht]
    \centering
    \scriptsize
    \resizebox{\textwidth}{!}{%
    \begin{tabular}{p{0.1\textwidth}p{0.7\textwidth}}
    \toprule
        \textbf{Gold} & \textcolor{cyan}{\texttt{<premise>}}She's been doing this for 30 years\textcolor{cyan}{\texttt{</premise>}} \\
        \midrule
        \textbf{ITFACD-OPT-6.7B} & \textcolor{cyan}{\texttt{<premise>}}She's been doing this \textbf{job} for 30 years\textcolor{cyan}{\texttt{</premise>}}\\
        \bottomrule
    \end{tabular}}
    \caption{Example of ACD with hallucination}
    \label{tab:hallucination-qualit-results}
\end{table}

A final, less frequent limitation of our approach concerns hallucination during generation. In rare cases, despite explicit instructions to preserve the original input verbatim and only insert XML tags, the models introduce new words, delete words, do paraphrasing, and, in some cases, generate short additional sub-sequences. This behavior alters the generated text and leads to span misalignment.
We experimented with several mitigation strategies, including lowering the temperature, restricting nucleus sampling, strengthening prompt constraints, and explicitly emphasizing verbatim reproduction in the instructions. However, these measures did not fully eliminate the issue. This suggests that the phenomenon is likely intrinsic to autoregressive generative models, which are optimized for fluent continuation rather than strict copying. An illustrative example of this phenomenon is provided in Table \ref{tab:hallucination-qualit-results}. While relatively rare, such hallucinations remain a key challenge for generative formulations of ACD and, more broadly, for structured prediction tasks with LLMs, highlighting the need for future work on constrained decoding strategies or hybrid extractive–generative approaches that better enforce input fidelity.

\subsection*{ACD examples with argument discovery and type refinement}
\begin{table}[ht]
    \centering
    \scriptsize
    \resizebox{\textwidth}{!}{%
    \begin{tabular}{p{0.1\textwidth}p{0.7\textwidth}}
    \toprule
        \textbf{Gold} & As a former educator, I believe that every child deserves equal access to education in the least restrictive environment possible. As a parent, I support that concept as long as the inclusion doesn't come at the expense of the other students in class. Inclusion can be successful rewarding for all involved but it also has drawbacks.\\
        \midrule
        
        \textbf{ITFACD-OPT-6.7B} & \textcolor{red}{\texttt{<claim>}}As a former educator,\textcolor{red}{\texttt{</claim>}} \textcolor{red}{\texttt{<claim>}}I believe that every child deserves equal access to education in the less restrictive environment possible.\textcolor{red}{\texttt{</claim>}}\textcolor{red}{\texttt{<claim>}}As a parent,\textcolor{red}{\texttt{</claim>}} \textcolor{cyan}{\texttt{<premise>}}I support that concept as long as the inclusion doesn't come at the expense of the other children in class.\textcolor{cyan}{\texttt{</premise>}}\textcolor{red}{\texttt{<claim>}}Inclusion can be successful rewarding for all involved but also has drawbacks.\textcolor{red}{\texttt{</claim>}}\\
        \bottomrule
    \end{tabular}}
    \caption{A second example of ACD with argument components discovery}
    \label{tab:discovery-1-qualit-results}
\end{table}

\begin{table}[ht]
    \centering
    \scriptsize
    \resizebox{\textwidth}{!}{%
    \begin{tabular}{p{0.1\textwidth}p{0.7\textwidth}}
    \toprule
        \textbf{Gold} & \textcolor{cyan}{\texttt{<premise>}}I'm not a politician\textcolor{cyan}{\texttt{</premise>}}, but \textcolor{red}{\texttt{<claim>}}I think I could go to Washington in a week and get everybody holding hands and get this bill signed because I talk to the Democratic leaders and they want it\textcolor{red}{\texttt{</claim>}}\\
        \midrule
        \textbf{ITFACD-OPT-6.7B} & \textcolor{cyan}{\texttt{<premise>}}I'm not a politician\textcolor{cyan}{\texttt{</premise>}}, but \textcolor{red}{\texttt{<claim>}}I think I could go to Washington in a week and get Everybody holding hands and get this Bill signed\textcolor{red}{\texttt{</claim>}} \textcolor{cyan}{\texttt{<premise>}}because I talk to the Democratic Leaders and they want it\textcolor{cyan}{\texttt{</premise>}}\\
        \bottomrule
    \end{tabular}}
    \caption{A second example of ACD with argument type refinement}
    \label{tab:type-enhancement-1-qualit-results}
\end{table}

\subsection*{Prompt design}
We provide, in Figure \ref{fig:eg_prompt_answer}, an illustration of our prompt used to instruction-tune our models.
\begin{figure*}[ht]
    \centering
    \includegraphics[width=.9\linewidth]{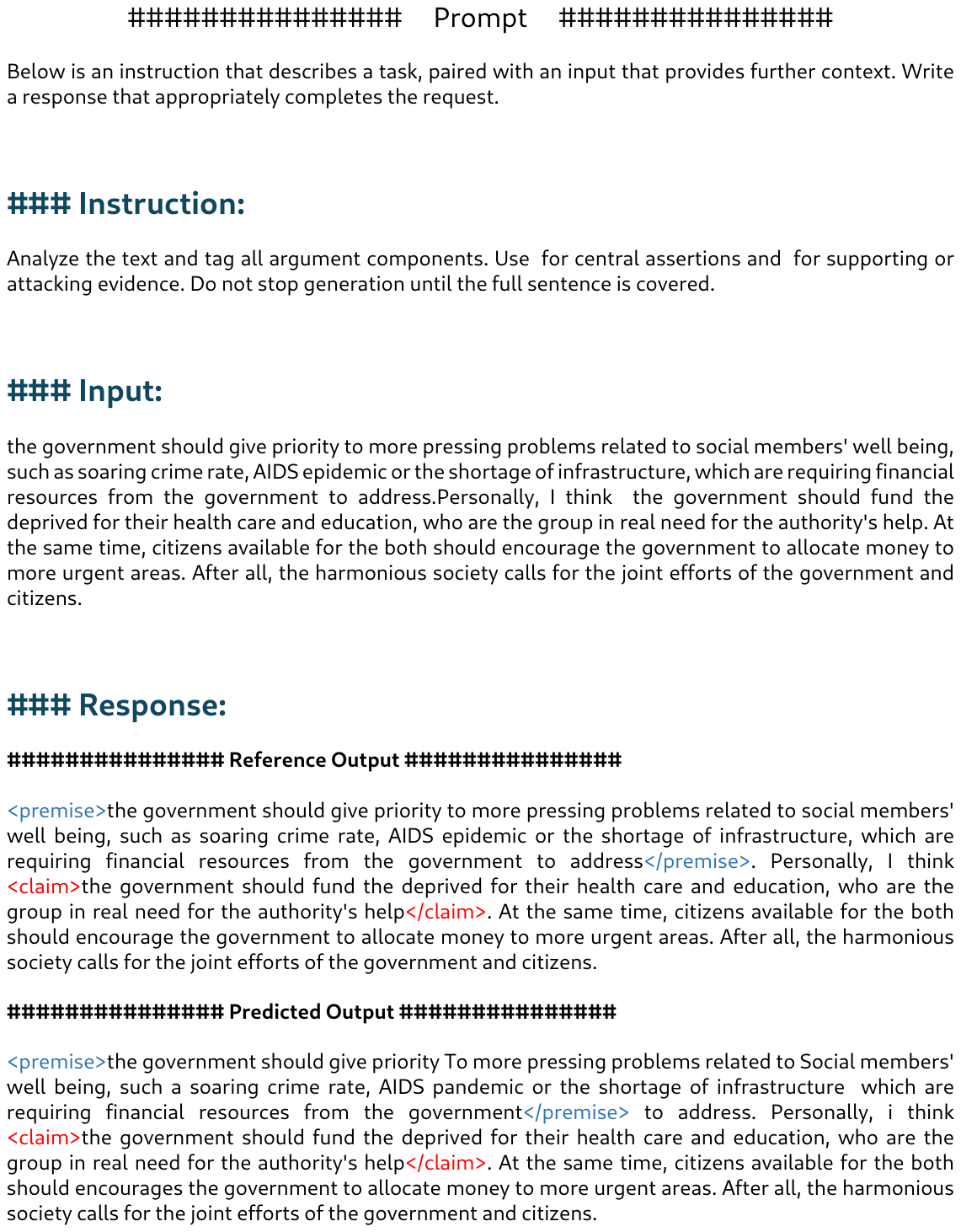}
    \caption{Full prompt and generation example for ACD task in AM}
    \label{fig:eg_prompt_answer}
\end{figure*} 

\end{document}